\documentclass[conference]{IEEEtran}
\IEEEoverridecommandlockouts

\usepackage{amsmath,amssymb,amsfonts}
\usepackage{array,multirow,color,booktabs,textcomp}
\usepackage{algorithmic}
\usepackage{float}
\usepackage{hyperref}
\usepackage{mathtools}
\usepackage{bm}
\usepackage{amssymb}
\usepackage{preview}
\usepackage{enumitem}
\usepackage{caption}
\usepackage{balance}
\usepackage{threeparttable,booktabs}
\usepackage{tipa}
\usepackage{algorithm}
\usepackage{algorithmic}
\usepackage{multirow}
\usepackage{subcaption}
\usepackage{cite}
\usepackage[table]{xcolor}

\usepackage{comment}
\usepackage{color, colortbl}
\def\BibTeX{{\rm B\kern-.05em{\sc i\kern-.025em b}\kern-.08em
    T\kern-.1667em\lower.7ex\hbox{E}\kern-.125emX}}
\definecolor{Gray}{gray}{0.9}

\begin{document}

\title{Is It Still Fair? Investigating Gender Fairness in Cross-Corpus Speech Emotion Recognition
}
\author{Shreya G. Upadhyay, Woan-Shiuan Chien, Chi-Chun Lee \\
    Department of Electrical Engineering, National Tsing Hua University, Taiwan. \\
    shreya@gapp.nthu.edu.tw, wschien@gapp.nthu.edu.tw, cclee@ee.nthu.edu.tw
}

\maketitle

\begin{abstract}
\emph{Speech emotion recognition} (SER) is a vital component in various everyday applications. Cross-corpus SER models are increasingly recognized for their ability to generalize performance. However, concerns arise regarding fairness across demographics in diverse corpora. Existing fairness research often focuses solely on corpus-specific fairness, neglecting its generalizability in cross-corpus scenarios. Our study focuses on this underexplored area, examining the gender fairness generalizability in cross-corpus SER scenarios. We emphasize that the performance of cross-corpus SER models and their fairness are two distinct considerations. Moreover, we propose the approach of a combined fairness adaptation mechanism to enhance gender fairness in the SER transfer learning tasks by addressing both source and target genders. Our findings bring one of the first insights into the generalizability of gender fairness in cross-corpus SER systems.
\end{abstract}


\section{Introduction}

\emph{Speech emotion recognition} (SER) has made significant strides by reshaping the landscape of human-computer interaction and enabling emotion-aware applications \cite{devillers2010real,10.5555/3295222.3295349, Acosta_2009}. Having more generalized SER models has led to growing interest in cross-corpus SER modeling, which aims at developing models capable of generalizing across diverse corpora and linguistic contexts. This entails many strategies to compensate features, domains, or label mismatches, using techniques such as transfer, semi-supervised, and few-shot learning \cite{parthasarathy2020semi, ahn2021cross,latif2022self,upadhyay2023phonetic}.  While these models have shown remarkable recognition performance on the target corpus but what about the fairness in performance?

In recent years, there has been an increasing emphasis on the trustworthiness of intelligent models, particularly concerning aspects such as privacy, fairness, safety, etc. Specifically, The notion of \emph{fairness} has gained substantial attention and become a focal point of extensive discussions. Numerous studies within the SER domain have recognized the issue of fairness, highlighting the model exhibits biases with sensitive attributes. To address these fairness concerns, various techniques have been proposed, including the use of adversarial networks, reweighing schemes, and the development of loss functions designed to mitigate biases \cite{coston2019fair, schmitz2022bias, chien2023achieving, maheshwari2023fairgrad}. Some studies have specifically aimed at neutralizing different attributes, such as gender or age, within SER approaches \cite{gorrostieta2019gender, chien2023achieving}. Most of these efforts have been focused on single-dataset (source-only) scenarios. However, the unfairness can manifest across multiple levels including sample, data distribution, modality, labeling, and domain-related aspects \cite{schmitz2022bias, chien2023achieving, gorrostieta2019gender}. The effectiveness of the model's fairness performance when deployed on another corpus has received limited, if any, exploration in the context of cross-domain SER scenarios. 
Emotions, with their intricate nature and cultural influences, manifest a diverse spectrum of expressions and interpretations across different domains and subjects. Hence, it becomes uncertain whether a SER model excelling in one emotional domain will demonstrate similar performance in another, particularly when accounting for cultural-linguistic variations.

However, a SER model tailored to a specific source corpus may introduce unfairness related to sensitive attributes like gender or age in the target corpus, amplifying some emotions while neglecting others. Cross-corpus SER models, designed for generalization across domains or cultural contexts \cite{costantini2022emotion, pastor2023cross}, must also ensure fairness across diverse demographics within these domains.
This study emphasizes the dual considerations of performance and fairness in cross-corpus SER models. We investigate the generalizability of fairness in cross-lingual SER models, employing a few state-of-the-art transfer learning and fairness techniques from the literature. Gender is specifically examined as the protected attribute of interest. Our study evaluates whether a cross-lingual SER model, which demonstrates fairness within its source corpus by showing no gender sensitivity, maintains this fairness when applied to the target corpus. 
This study introduces a novel perspective in cross-corpus SER, underscoring the importance of integrating fairness considerations alongside performance when deploying SER systems across corpora.

In this study, for cross-corpus SER fairness investigations, we utilize two large naturalistic speech corpora, MSP-Podcast \cite{lotfian2017building} and BIIC-Podcast \cite{upadhyay2023an}. 
Our experimental results reveal two insights, first despite exhibiting source-specific fairness with various transfer learning approaches, cross-corpus SER generalization can introduce gender biases for the target corpus which questions the fairness of the SER model. Second, we propose a simple fairness mechanism of combined fairness adaptation that integrates fairness towards the source and the target corpus while modeling a cross-corpus SER. Our proposed cross-corpus setting based \emph{Combined Fairness Adaptation} (\emph{CFA}) idea achieves significantly better gender fairness (GF) compared to the considered baselines of this study. The preliminary findings of this study demonstrate the effectiveness and necessity of research in this direction.

\newcolumntype{g}{>{\columncolor{Gray}}c}
\begin{table*}[]
\centering
\caption{The gender-specific performances (in UAR) and their fairness (\(\Delta SP\) and \(\Delta EO\)) of the models in fairness evaluations for each considered emotion; tests on MSP-P (white rows) and BIIC-P (gray rows) for the cross-corpus SER model.}
\vspace{-2mm} 
\renewcommand{\arraystretch}{1.0}
\resizebox{1\textwidth}{!}{
\begin{tabular}{c|gg|gg|gg|gg||gg|gg|gg|gg}
\toprule\specialrule{\cmidrulewidth}{0pt}{0pt}
   & \multicolumn{8}{c||}{\textbf{Gender-Specific Performance}}                                                             & \multicolumn{8}{c}{\textbf{Fairness}}                                                                                   \\ \hline \rowcolor{white}
    & \multicolumn{2}{c|}{Neutral}       & \multicolumn{2}{c|}{Happiness} & \multicolumn{2}{c|}{Anger} & \multicolumn{2}{c||}{Sadness} & \multicolumn{2}{c|}{Neutral}    & \multicolumn{2}{c|}{Happiness}   & \multicolumn{2}{c|}{Anger}   & \multicolumn{2}{c}{Sadness}   \\\hline \rowcolor{white}
    & M     & F     & M          & F          & M          & F          & M          & F          & $\Delta$ SP          & $\Delta$ EO          &  $\Delta$ SP          & $\Delta$ EO          & $\Delta$ SP          & $\Delta$ EO          & $\Delta$ SP          & $\Delta$ EO          \\ \hline \hline \rowcolor{white}
    \multirow{2}{*}{FS \cite{ahn2021cross}}  & 67.08 & 69.66 & 66.69 & 70.00 & 74.22 & 70.94 & 70.54 & 65.58 & 0.305 & 0.215 & 0.385 & 0.355 & 0.343 & 0.335 & 0.339 & 0.423 \\ \cline{2-17} 
 &58.44 & 48.33 & 51.93 & 64.47 & 57.74 & 70.42 & 51.69 & 62.22 & 0.528 & 0.625 & 0.512 & 0.624 & 0.558 & 0.599 & 0.503 & 0.435 \\ \hline\rowcolor{white}
 \multirow{2}{*}{GAN \cite{su2022unsupervised}}   & 61.47 & 67.67 & 57.23 & 62.73 & 69.50 & 70.30 & 59.45 & 59.92 & 0.348 & 0.302 & 0.386 & 0.304 & 0.306 & 0.329 & 0.304 & 0.396 \\ \cline{2-17} 
 & 58.33 & 44.44 & 64.39 & 47.57 & 63.72 & 47.86 & 46.08 & 66.04 & 0.570 & 0.667 & 0.559 & 0.632 & 0.436 & 0.465 & 0.417 & 0.512 \\ \hline\rowcolor{white}
 \multirow{2}{*}{PA \cite{upadhyay2023phonetic}}   & 75.22 & 73.16 & 64.01 & 68.85 & 65.41 & 68.29 & 60.36 & 65.06 & \textbf{0.352 } & \textbf{0.341 } & \textbf{0.375 } & \textbf{0.351 } & \textbf{0.380 } & \textbf{0.384 } & \textbf{0.356 } & \textbf{0.322 } \\ \cline{2-17} 
 & 65.11 & 57.99 & 63.58 & 57.75 & 58.01 & 71.79 & 63.57 & 52.51 & \textbf{0.541 } & \textbf{0.515 } & \textbf{0.540 } & \textbf{0.565 } & \textbf{0.534 } & \textbf{0.459 } & \textbf{0.567 } & \textbf{0.549 } \\ \hline\rowcolor{white}
\multirow{2}{*}{PA-FairW \cite{chakraborty2020fairway}} & 73.17 & 79.42 & 74.37 & 75.73 & 69.88 & 60.12 & 74.02 & 63.88 & 0.133 & 0.253 & 0.195 & 0.205 & 0.279 & 0.242 & 0.253 & 0.221 \\ \cline{2-17} 
 & 69.43 & 57.37 & 59.72 & 69.42 & 48.23 & 64.09 & 65.33 & 52.26 & 0.346 & 0.476 & 0.462 & 0.498 & 0.335 & 0.428 & 0.412 & 0.511 \\ \hline\rowcolor{white}
 \multirow{2}{*}{PA-ReW \cite{kamiran2012data}}  & 73.25 & 74.05 & 73.91 & 70.43 & 59.93 & 63.39 & 65.63 & 68.01 & \textbf{0.121} & \textbf{0.256} & \textbf{0.135} & \textbf{0.194} & \textbf{0.159 } & \textbf{0.168}  & \textbf{0.277 } & \textbf{0.224 } \\ \cline{2-17} 
 & 67.25 & 59.05 & 58.91 & 69.43 & 49.93 & 58.39 & 54.63 & 65.01 & \textbf{0.320 } & 0\textbf{.401 } & \textbf{0.411}  & \textbf{0.512 } & \textbf{0.321}  & \textbf{0.416 } & \textbf{0.404}  & \textbf{0.415 } \\
\specialrule{\cmidrulewidth}{0pt}{0pt}\bottomrule                         
\end{tabular}}
\vspace{-0.2cm}
\label{tab:SER_fairness_analysis}
\end{table*}

\newpage
\section{Fairness Generalizability Analyses}
\label{sec:fair_analysis}

\subsection{Naturalistic Corpora}

\noindent
\textbf{The MSP-Podcast (MSP-P)} \cite{lotfian2017building} corpus contains 166 hours of emotional \emph{American English} speech (v1.10), sourced from audio-sharing websites. This resource is valuable for SER research due to its extensive size and emotionally balanced dialogues from different individuals. It includes annotations for primary and secondary emotions and emotional attributes. We select 49,018 samples from this corpus, evenly divided between 24,466 male and 24,552 female subject samples. 

\smallskip
\noindent
\textbf{The BIIC-Podcast (BIIC-P)} \cite{upadhyay2023an} corpus is a SER database (v1.0) in \emph{Taiwanese Mandarin}. It contains 157 hours of speech samples from podcasts and follows a data collection methodology similar to the MSP-P corpus. The annotations cover primary and secondary emotional categories, as well as three emotional attributes. Here, we utilize 18,706 samples of the data with 9,654 male and 9,326 female subject samples.

\smallskip
\noindent
This work considers four primary emotion categories (\emph{Neutral, Happiness, Anger}, and \emph{Sadness}) samples for binary SER tasks with predefined train-valid-test sets of the corpora. For a detailed analysis, we analyze each emotion individually rather than as a 4-category SER task.

\subsection{Experimental Settings and Evaluation Metrics}

\label{settings}
In our experiments, we use MSP-P as the emotion-labeled source corpus and BIIC-P as the target corpus. We employ popular wav2vec2.0 \cite{baevski2020wav2vec} feature vectors along with different transfer learning and fairness considering architectures from the literature to investigate fairness generalizability on the target corpus in cross-corpus SER models. Optimization is performed using the Adam optimizer with a learning rate of 0.0001 and a decay factor of 0.001, while training involves back-propagation with binary cross-entropy loss. Training proceeds for a maximum of 50 epochs with a batch size of 64 and includes early stopping. Evaluation of SER model performance is based on the Unweighted Average Recall (UAR). To measure fairness, we utilize established metrics: Statistical Parity (\(\Delta SP\)) and Equal Odds (\(\Delta EO\)). This ranges from -1 to 1, where values closer to 0 indicate better fairness. These metrics are commonly used in fairness studies \cite{schmitz2022bias, bellamy2018ai}. 
\(\Delta SP\) ensures the same proportion of positive outcomes across groups, regardless of their true characteristics (e.g., gender, race). \(\Delta EO\) requires the model to have equal true positive and false positive rates for different groups, ensuring fair performance. By analysing both we are making more consistent investigation.

\vspace{-0.06cm}
\subsection{Fairness Generalizability Evaluations}
\label{sec:fairness}
To examine the generalizability of \emph{gender fairness} (GF) in cross-corpus SER models, we perform cross-test fairness assessments using state-of-the-art (SOTA) methods in two stages. First, we evaluate several cross-corpus techniques using effective SOTA transfer learning (TL) methods to assess GF generalizability. Second, we apply fairness approaches from the literature to the best-performing TL model to assess GF generalizability after incorporating a source-specific fairness approach. Table \ref{tab:SER_fairness_analysis} summarizes SER performances (UAR) and fairness evaluations (\(\Delta SP\) and \(\Delta EO\)) across four primary emotions for male and female speakers. Results are averaged from ten experiment repetitions. To assess whether the cross-corpus model that demonstrates GF on the source also maintains GF on the target corpus (beyond just performance), Table \ref{tab:SER_fairness_analysis} presents the results for both the MSP-P and BIIC-P test sets for the respective cross-corpus model.

\smallskip
\noindent
\emph{\textbf{GF Generalizability with Various TL Approaches:}}
To analyze GF generalizability using TL approaches, we consider three SOTA methods: Few-shot (\emph{FS}) \cite{ahn2021cross}, which leverages knowledge from source corpora and adapts the model to the target domain, GAN-based (\emph{GAN}) \cite{su2022unsupervised}, which employs adversarial training, and phonetically-anchored (\emph{PA}) \cite{upadhyay2023phonetic}, which utilizes learning in a shared phonetic space for SER models. Table \ref{tab:SER_fairness_analysis} highlights the gender-specific performance of these TL models, revealing disparities between males and females over the target BIIC-P corpus. For instance, in the \emph{Anger} category, UARs for males are 57.74\%, 63.72\%, and 58.01\%, and for females, 70.42\%, 47.86\%, and 71.79\% for FS, GAN, and PA models, respectively. This performance discrepancy is also evident in the source MSP-P test set. Fairness metrics in Table \ref{tab:SER_fairness_analysis} show gender unfairness across emotional classes over both corpora test sets. For example, in \emph{Anger}, the \(\Delta SP\) values are 0.380 and 0.534 for PA models for MSP-P and BIIC-P, respectively. This behavior can be seen in other emotion categories as well. The results highlight the challenges in GF generalizability for cross-corpus SER TL models, emphasizing the need to address fairness in such scenarios distinctly.

\smallskip
\noindent
\emph{\textbf{Cross-Corpus GF Generalizability with Fairness Techniques:}}
Given our emphasis on GF in cross-corpus SER settings, we consider two state-of-the-art fairness methods from the literature: \emph{Fairway} \cite{chakraborty2020fairway}, \emph{Reweigh} \cite{kamiran2012data}. These fairness methodologies are source-specific. However, existing literature lacks fairness techniques specifically designed for cross-corpus settings. Therefore, we proceed to test these existing fairness techniques in conjunction with our best-performing TL method (\emph{PA}) from the previous section.
Table \ref{tab:performance} shows the results of \emph{PA-FairW} (\emph{PA} +\emph{Fairway}) and \emph{PA-ReW} (\emph{PA} + \emph{Reweigh}).
Upon analyzing the performance of these models over MSP-P and BIIC-P test sets, we observe that although there is a decrease in the \(\Delta SP\) and \(\Delta EO\) values for MSP-P (indicating improvement), but no significant reduction is observed for BIIC-P. For instance, for \emph{Anger}, for MSP-P, (\emph{PA-ReW}) yields \(\Delta SP\) and \(\Delta EO\) values of 0.159 and 0.168, respectively, while on BIIC-P set, the \(\Delta SP\) and \(\Delta EO\) values are still high with 0.321 and 0.416, respectively (shows no improvement). Similar behavior is observed in other emotion categories as well. These findings indicate that even if the model exhibits better source-specific GF, these approaches fail to generalize GF across cross-evaluations.

\vspace{-0.1cm}
\section{Fair Cross-Corpus SER}
\label{sec:modeling}

Building on insights from Section \ref{sec:fair_analysis}, cross-corpus SER models face challenges with GF generalizability in target corpora. To address this, we propose a strategy (Fig. \ref{fig:arch}) for the GF generalized cross-corpus SER. The architecture has two key blocks: an emotion classification (\emph{EC}) block and a combined fairness adaptation (\emph{CFA}) block. For testing our idea, in the \emph{EC} block, we use a basic SER architecture; processing wav2vec2.0 \cite{baevski2020wav2vec} features with a transformer and four fully-connected layers for binary emotion classification. The \emph{CFA} block adds an auxiliary adversarial network for gender classification (\emph{GC}) with a reverse gradient layer which transforms \emph{EC} features into a gender-neutral state by minimizing \emph{GC}'s ability to predict gender from both corpora. Other experimental settings and evaluation metrics match those in Section \ref{settings}.

The training process unfolds in stages. Initially, we train the primary SER model on the source corpus, focusing on capturing gender-neutral emotional features while disregarding gender-specific information. Subsequently, we create mini-batches that combine data from both the MSP-P and BIIC-P for gender-neutral adversarial training. The \emph{EC} aims to generate gender-neutral embeddings, while the \emph{GC} aims to make accurate gender predictions. This training process uses a binary cross-entropy loss function for classifying gender, penalizing the classifier for making gender predictions based on shared feature representations. In each training iteration, we update both the primary model's weights and the gender classifier's parameters. The inserted reverse gradient layer in learning modifies the gradients backpropagated from the \emph{GC} branch to make the shared feature representations more gender-neutral. Equation \ref{eq1} shows the binary classification loss used for \emph{EC} and \emph{GC} tasks.
{\setlength\abovedisplayskip{1.5mm}
\setlength\belowdisplayskip{1.5mm}
\begin{equation}
\label{eq1}
\resizebox{.91\hsize}{!}{$
{L}(\theta_{D})  = -\frac{1}{N}\sum_{i=1}^{N}\left[ y_{i}\log P(y=1|z_{i})+(1-y_{i}) \log P(y=0|z_{i})\right]
$}
\end{equation}

where \(z\) is the intermediate embeddings produced by the primary \emph{EC} model. \(y\) be the true labels for emotions (0 for target, 1 for other) for \emph{EC} and the gender of speakers (0 for male, 1 for female) for \emph{GC} task. \(D\) represents the \emph{EC} or \emph{GC}, \(P(y|z)\) be the predicted target attribute probability distribution given \(z\), and \(\theta_{D}\) be the parameters. 
To enforce a similarity between source and target genders, we use a contrastive loss that encourages samples with the same gender to be close in the feature space and those with different genders to be distant from each other. The goal is to make the learned representations of gender-related features similar between the source and target domains. The Equation \ref{eq2} is used to integrate a source and target similarity loss.
{\setlength\abovedisplayskip{1.5mm}
\setlength\belowdisplayskip{1.5mm}
\begin{equation}
\label{eq2}
  L_{GSim}(x_1, x_2, y)  = (1-y)\frac{1}{2}D^2 + y\frac{1}{2}max(0, m-D)^2
\end{equation}}
where D is the Euclidean distance between the embeddings of samples \(x_1\) and \(x_2\). m is the margin and here is fixed to 1. This loss encourages the model to minimize the distance between samples with the same gender (\(y=0\)) and maximize the distance between samples with different genders (\(y=1\)).
Equation \ref{eq3} illustrates the total loss for the \emph{PA-CFA} model, which combines the \emph{EC} loss and the \emph{GC} loss.
{\setlength\abovedisplayskip{2mm}
\setlength\belowdisplayskip{2mm}
\begin{equation}
\label{eq3}
  L_{total}  = {L}_{EC} + \alpha * {L}_{GSim}- \beta * {L}_{GC} 
\end{equation}}
where \({L}_{emo}\) and \({L}_{GC}\) are the binary emotion classification and gender classification loss. \({L}_{GSim}\) is the loss for the gender similarity. \(\alpha\) and \(\beta\) are set to 0.5. Here, during training, target gender labels are needed, but not during inference.

\begin{figure}[tbp]
  \centering
  \includegraphics[height=0.533\linewidth,width=\linewidth]{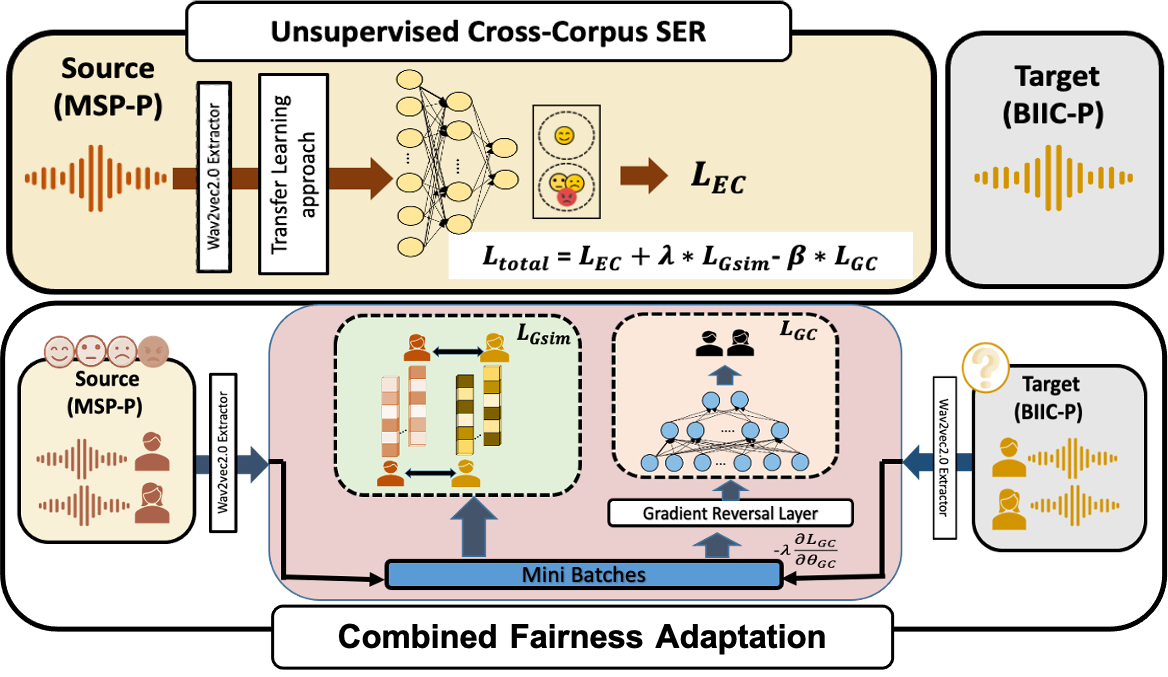}
  \caption{Proposed \emph(CFA) SER approach using  gender-neutral adaptation mechanism for fair cross-corpus SER.}
  \label{fig:arch}
  \vspace{-0.4cm}
\end{figure}

\newcolumntype{w}{>{\columncolor{white}}c}
\newcolumntype{g}{>{\columncolor{Gray}}c}
\begin{table*}[]
\centering
\caption{The gender-specific performances (in UAR) and fairness (in $\Delta SP$, $\Delta EO$) for each SER task with the test over both MSP-P (in White color rows) and BIIC-P (in Gray color rows) test sets. It includes the statistical test over PA-ReW and the proposed (CFA) model fairness values, denoted by asterisks (* for p \(<\) 0.1, ** for p \(<\) 0.05).}
\vspace{-2mm}
\renewcommand{\arraystretch}{1.1}
\resizebox{1\textwidth}{!}{%
\begin{tabular}{c|gg|gg|gg|gg||gg|gg|gg|gg}
\toprule\specialrule{\cmidrulewidth}{0pt}{0pt}
   & \multicolumn{8}{c||}{\textbf{Gender-Specific Performance}}                                                             & \multicolumn{8}{c}{\textbf{Fairness}}                                                                                   \\ \hline \rowcolor{white}
    & \multicolumn{2}{c|}{Neutral}       & \multicolumn{2}{c|}{Happiness} & \multicolumn{2}{c|}{Anger} & \multicolumn{2}{c||}{Sadness} & \multicolumn{2}{c|}{Neutral}    & \multicolumn{2}{c|}{Happiness}   & \multicolumn{2}{c|}{Anger}   & \multicolumn{2}{c}{Sadness}   \\\hline \rowcolor{white}
     & M & F & M & F & M & F & M & F & $\Delta$SP & $\Delta$EO & $\Delta$SP & $\Delta$EO & $\Delta$SP & $\Delta$EO & $\Delta$SP & $\Delta$EO \\ \hline \hline \rowcolor{white}
\multirow{2}{*}{{PA-ReW}} & 72.63 & 75.96  & 68.82 & 67.63 & 68.42 & 70.05& 63.67 & 68.14 & 0.161 & 0.159 & 0.198 & 0.257 & 0.111 & 0.203 & 0.154 & 0.248 \\ \cline{2-17} 
&59.03 & 68.44 & 59.31 & 64.57  & 52.98 & 45.37  & 61.81 & 53.03 & 0.391 & 0.401 & 0.412 & 0.469 & 0.363 & 0.423 & 0.398 & 0.402\\ \toprule\specialrule{\cmidrulewidth}{0pt}{0pt} 
\rowcolor{white}
\multirow{2}{*}{{Base-CFA}}  & 73.25 & 70.74  & 75.30 & 71.02  & 75.33 & 76.32  & 71.18 & 73.70& 0.256 & 0.264 & 0.190 & 0.164 & 0.130 & 0.153 & 0.132 & 0.133 \\ \cline{2-17} 
 & 67.60 & 64.88 & 63.05 & 62.40  & 69.28  & 71.56 & 64.94 & 60.86 & 0.335\small{*}  & 0.294\small{**}  & 0.212\small{**}  & 0.263\small{**}  & 0.256\small{**}  & 0.205\small{**}  & 0.193\small{**}  & 0.275\small{**} \\\hline
 \rowcolor{white}
\multirow{2}{*}{\textbf{{PA-CFA}}}   & 74.44 & 71.33 & 76.93 & 70.47 & 79.74 & 82.42  & 70.69 & 74.32 & 0.211 & 0.231 & 0.119 & 0.216 & 0.107 & 0.195 & 0.106 & 0.201 \\  \cline{2-17} 
& 68.24 & 65.75  & 61.55 & 64.83 & 68.15  & 70.14  & 65.36 & 62.35 & \textbf{0.205}\small{**}  & \textbf{0.211}\small{**}  & \textbf{0.223 }\small{**} & \textbf{0.206}\small{**}  & \textbf{0.260}\small{**}  & \textbf{0.287}\small{**}  & 0.\textbf{236}\small{**}  & \textbf{0.241}\small{**} \\ \toprule\specialrule{\cmidrulewidth}{0pt}{0pt}
\rowcolor{white}
 \multirow{2}{*}{{PA-Adv}}   & 74.63 & 71.91 & 75.04 & 71.06 & 72.61 & 76.08 & 69.99 & 72.35 & 0.295 & 0.301 & 0.21 & 0.234 & 0.192 & 0.205 & 0.184 & 0.217 \\  \cline{2-17} 
 & 62.89 & 66.05 & 63.94 & 66.38  & 70.18 & 68.84  & 61.23 & 64.21 & 0.281 & 0.345 & 0.273 & 0.357 & 0.322 & 0.306 & 0.251 & 0.365 \\ 
 \specialrule{\cmidrulewidth}{0pt}{0pt}\bottomrule             
\end{tabular}}
\vspace{-0.1cm}
\label{tab:performance}
\end{table*}

To align genders between source and target domains, we apply a contrastive loss to encourage similar gender samples to be closer in feature space and dissimilar ones to be farther apart. Equation \ref{eq2} illustrates how we integrate this source and target similarity loss.
{\setlength\abovedisplayskip{1.5mm}
\setlength\belowdisplayskip{1.5mm}
\begin{equation}
\label{eq2}
  L_{GSim}(x_1, x_2, y)  = (1-y)\frac{1}{2}D^2 + y\frac{1}{2}max(0, m-D)^2
\end{equation}}
where D is the Euclidean distance between the embeddings of samples \(x_1\) and \(x_2\). m is the margin and here is fixed to 1. This loss encourages the model to minimize the distance between samples with the same gender (\(y=0\)) and maximize the distance between samples with different genders (\(y=1\)).
Equation \ref{eq3} illustrates the total loss for the \emph{PA-CFA} model, which combines the \emph{EC} loss and the \emph{GC} loss.
{\setlength\abovedisplayskip{2mm}
\setlength\belowdisplayskip{2mm}
\begin{equation}
\label{eq3}
  L_{total}  = {L}_{EC} + \alpha * {L}_{GSim}- \beta * {L}_{GC} 
\end{equation}}
where \({L}_{emo}\) and \({L}_{GC}\) are the binary emotion classification and gender classification loss. \({L}_{GSim}\) is the loss for the gender similarity. \(\alpha\) and \(\beta\) are set to 0.5. Here, during training, target gender labels are needed, but not during inference.

\vspace{-0.1cm}
\section{Experiment and Analyses}

Table \ref{tab:performance} shows gender-specific performance and fairness metrics for \emph{Base-CFA} with a basic SER model and \emph{PA-CFA}, combined with the \emph{PA} TL model.
Table \ref{tab:performance} demonstrates that on cross-evaluation (on BIIC-P test set), our proposed \emph{CFA} baseline \emph{Base-CFA} method outperforms \emph{PA-ReW} in terms of GF generalizability across most emotion categories. For example,  for \emph{Anger}, \emph{Base-CFA} achieves balanced gender-specific performance  (M:69.28\%, F:71.56\%) compared to \emph{PA-ReW} (M:52.98\%, F:45.37\%). In terms of fairness metrics, \emph{CFA} shows a significant reduction in both  \(\Delta SP\) and \(\Delta EO\) across all emotions compared to the \emph{PA-ReW}. For instance, the SP value for \emph{Anger} decreases by 0.107, reflecting improved GF. Additionally, \emph{PA-CFA}, which combines phonetic-based constraints with CFA, captures language-specific information in cross-lingual tasks, exceeds the performance of \emph{PA-ReW}. For example, for Anger,\(\Delta SP\) drops from 0.363 to 0.260, a consistent trend across both fairness metrics.

We validate our approach by training a model without the gender similarity loss (\emph{\({L}_{GSim}\)}), called \emph{PA-Adv}. Table \ref{tab:performance} shows that \emph{PA-CFA} still achieves better fairness. For example, in \emph{Anger}, \emph{PA-Adv} has a \(\Delta SP\) of 0.322, while \emph{PA-CFA} achieves 0.260, highlighting the effectiveness of \emph{\({L}_{GSim}\)} in improving fairness by aligning gender features across corpora. For reference, Table \ref{tab:UAR} presents the overall UAR performance of the \emph{CFA} model. From Table \ref{tab:UAR}, we can see that \emph{PA-CFA} shows a slight decrease in performance compared to \emph{PA}, but it remains competitive over all emotion categories. We believe this minor performance drop is a worthwhile trade-off for improved GF.  For reference, the overall UAR represents recall across all classes, irrespective of gender. In Table \ref{tab:UAR}, UAR values are shown separately for males and females, so the overall UAR shown in Table \ref{tab:UAR} is not an average of gender-specific UARs. 

\smallskip
\noindent
\textbf{Analysis:}
We utilize t-SNE plots to explore gender attribute feature spaces in \emph{PA-CFA} and \emph{PA-ReW}, shown in Figure \ref{fig:feature_space}, aiming to understand why \emph{PA-CFA} performs better. In Figure \ref{fig:cl-hap}, comparing \emph{Anger} in Figure \ref{fig:cl-hap} to Figure \ref{fig:clg-hap}, we see distinct male-female clusters in \ref{fig:cl-hap} and a mixed distribution in \ref{fig:clg-hap}. This indicates coexisting male and female features from different corpora, across all emotions studied. It underscores the importance of integrating \emph{CFA} and questions the effectiveness of source-specific fairness in cross-corpus SER. 

\begin{table}[tbp]
\centering
\caption{Overall UAR performances on BIIC-P.}
\vspace{-2mm}
\renewcommand{\arraystretch}{1.1}
\resizebox{0.8\columnwidth}{!}{%
\begin{tabular}{c|cccc}
\toprule\specialrule{\cmidrulewidth}{0pt}{0pt}
 & Neutral &Happiness &Anger  &Sadness\\ \hline \hline
PA & 74.53 & 74.87 & 76.46 & 72.37\\
PA-CFA & 72.97  & 72.37 & 75.30 & 70.82\\
\specialrule{\cmidrulewidth}{0pt}{0pt}\bottomrule                          
\end{tabular}}
\label{tab:UAR}
\vspace{-0.1cm}
\end{table}

We also analyze embeddings from both the \emph{PA-ReW} and proposed \emph{PA-CFA} models to assess gender-related information across the MSP-P and BIIC-P contexts. Figure \ref{fig:GC} presents the gender detection (GD) accuracy, with MSP-P on the left and BIIC-P on the right. \emph{PA-ReW} exhibits varying GD accuracy: lower for MSP-P and higher for BIIC-P, suggesting residual gender information specific to BIIC-P despite lower MSP-P performance. Conversely, \emph{PA-CFA} shows improved gender neutralization, with similar GD accuracy across both corpora (e.g., for \emph{Anger}, 36\% MSP-P, 35\% BIIC-P in Figure \ref{fig:biic_GC}).
In \emph{PA-CFA}, there are improvements in emotions like \emph{Neutral} and \emph{Anger} for MSP-P GD accuracy compared to \emph{PA-ReW}, reflecting compensation in MSP-P performance due to enhanced gender fairness across MSP-P and BIIC-P corpora.

\begin{figure}[t!]    
\centering
    \begin{subfigure}{.35\textwidth}
       \includegraphics[width=\textwidth]{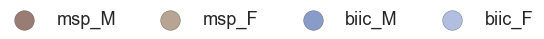}
    \end{subfigure}%
    
    \begin{subfigure}{.24\textwidth}
    \centering
        \includegraphics[width=0.85\textwidth, height=0.8\textwidth]{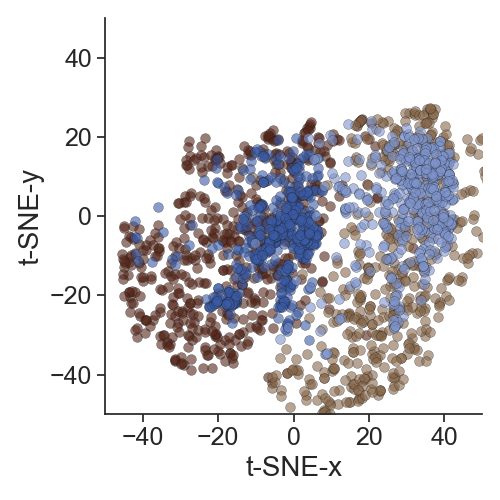}
        \subcaption{\emph{PA-ReW}}
        \label{fig:cl-hap}
    \end{subfigure}%
    \begin{subfigure}{.24\textwidth}
       \centering
       \includegraphics[width=0.85\textwidth, height=0.8\textwidth]{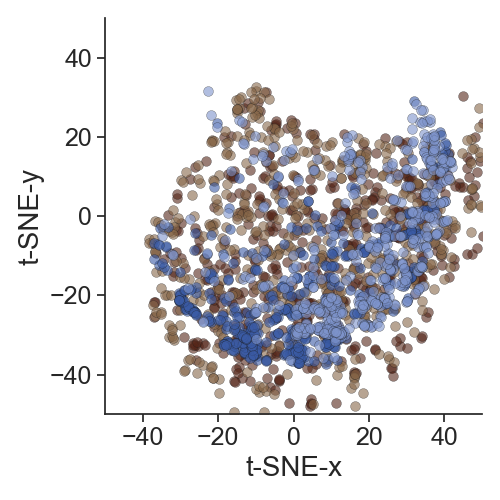}
       \subcaption{\emph{PA-CFA}}
       \label{fig:clg-hap}
    \end{subfigure}
    \caption{t-SNE plot for \emph{Anger} features.}
\label{fig:feature_space}
\vspace{-0.1cm}
\end{figure}

\begin{figure}[t!]
    \centering
    \begin{subfigure}{0.4\textwidth}
       \includegraphics[width=\textwidth]{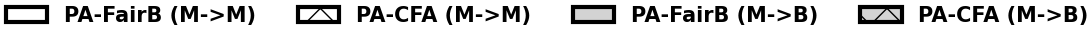}
    \end{subfigure}
    
     \begin{subfigure}{0.23\textwidth}
       \includegraphics[width=\textwidth]{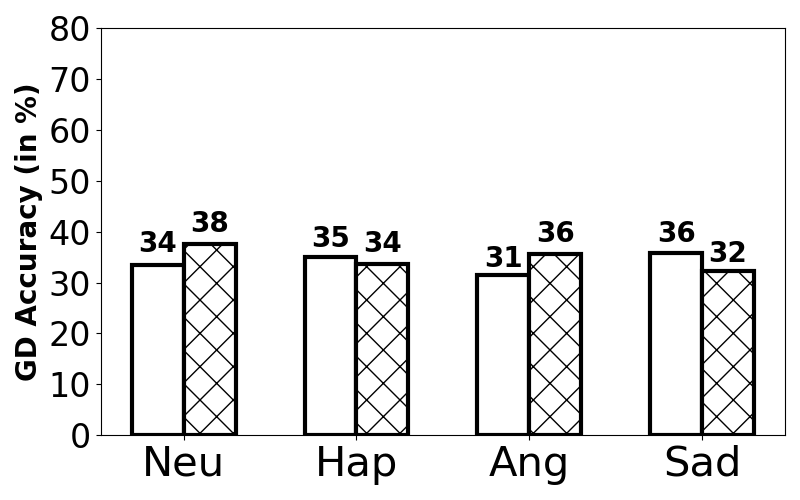}
       \subcaption{MSP-P}
       \label{fig:msp_GC}
    \end{subfigure}
    \begin{subfigure}{0.23\textwidth}
       \includegraphics[width=\textwidth]{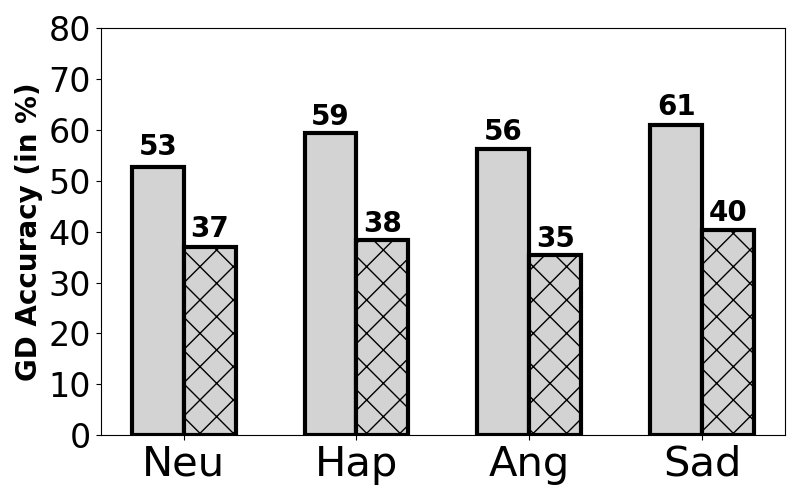}
       \subcaption{BIIC-P}
       \label{fig:biic_GC}
    \end{subfigure}
    \caption{Gender detection (GD) accuracy plot using PA-ReW and PA-CFA model embeddings.}
\label{fig:GC}    
\vspace{-0.3cm}
\end{figure}

\section{Conclusion}

\label{sec:Conclusion}


This research addresses overlooked fairness concerns in cross-corpus SER, particularly focusing on gender neutrality. Our findings reveal that cross-corpus SER models, while fair within their source corpus, introduce biases when generalized across different corpora. We propose an initial approach, \emph{combined fairness adaptation (CFA)}, to enhance gender neutrality across both source and target corpora in emotion transfer tasks. Initial experiments demonstrate the efficacy of our approach in creating gender-fair cross-corpus SER systems. Future research will refine our fairness mechanism through feature-side analysis to pinpoint specific areas where fairness issues arise in cross-corpus SER settings.

\newpage

\bibliographystyle{IEEEtran}
\bibliography{refs}

\end{document}